# Research on fusing topological data analysis with convolutional neural network

Yang Han.Qin Guangjun*.Liu Ziyuan.Hu Yongqing.Liu Guangnan.Dai Qinglong

Smart City College, Beijing Union University, Beijing

Qin Guangjun*,15652448798,zhtguangjun@buu.edu.cn

**Abstract:** Convolutional Neural Network (CNN) struggle to capture the multi-dimensional structural information of complex high-dimensional data, which limits their feature learning capability. This paper proposes a feature fusion method based on Topological Data Analysis (TDA) and CNN, named TDA-CNN. This method combines numerical distribution features captured by CNN with topological structure features captured by TDA to improve the feature learning and representation ability of CNN. TDA-CNN divides feature extraction into a CNN channel and a TDA channel. CNN channel extracts numerical distribution features, and the TDA channel extracts topological structure features. The two types of features are fused to form a combined feature representation, with the importance weights of each feature adaptively learned through an attention mechanism. Experimental validation on datasets such as Intel Image, Gender Images, and Chinese Calligraphy Styles by Calligraphers demonstrates that TDA-CNN improves the performance of VGG16, DenseNet121, and GoogleNet networks by 17.5%, 7.11%, and 4.45%, respectively. TDA-CNN demonstrates improved feature clustering and the ability to recognize important features. This effectively enhances the model's decision-making ability.

## 1. Introduction

CNN is an essential implementation of deep learning. The core idea behind CNN is to gradually learn abstract feature representations of data by applying multilevel nonlinear transformations and feature extraction. CNN can effectively capture local and global features, such as edges, corners, and textures in images[1–3]. They are a significant area of research in both the scientific community and industry.

CNN is proficient at processing data in Euclidean space but has difficulty capturing and utilizing multidimensional structural information when handling high-dimensional, complex structured data in non-Euclidean space[4]. This issue makes it difficult for CNN to effectively capture the intrinsic topological structure of data, thus failing to reveal and utilize the complex causal and structural relationships contained in data[5]. However, the deep structural information of data is crucial for understanding its intrinsic properties[6]. This information is often more revealing, highlighting structural relationships between components in specific data types. In recent years, models like Graph Neural Network (GNN) and Geometric Deep Learning (GDL) have attempted to extend neural networks to non-Euclidean spaces, such as graphs, manifolds, grids, and strings.[7]. Learning structural information from these models can help improve neural networks' feature representation and decision-making capabilities. TDA has also been gradually integrated into neural networks as an algebraic topology-based data analysis method. It can effectively extract deeper topological structure information from data and has shown excellent results.

TDA[8] can be deemed an unsupervised learning method based on the principles of topology[9]. It involves converting the data into a point cloud and utilizing techniques such as Persistent Homology[10-11] and Mapper[12-14] to transform high-dimensional data into a topological space for analyzing the shapes and relationships of the topological structures in point-cloud data. The method can effectively uncover the underlying correlations within data. The topological invariance property of these structures ensures that the extracted topological features remain robust when the data changes such as distortions, deformations, or the presence of a small amount of noise[15]. This property allows the topological features extracted by TDA to capture the essential characteristics of the data in complex data structures, surpassing the shallow understanding provided by other data analysis methods that focus solely on numerical features[16].

This paper proposes a method for conbining TDA with CNN and constructs a TDA-CNN model. The aim is to improve the feature learning and representation capability of CNN. The main contributions of this paper are as follows:

1. A fusion method of topological structural features and numerical features is proposed.

The method divides the feature extraction into two channels. The CNN channel extracts numerical distribution features, while the TDA channel extracts data structure features. The two features are fused into a combination of features according to the channel, aiming to improve the model's feature learning and representation ability.

2. Optimizing the conversion process from the Persistence Diagram (PD) to the Persistence Image (PI)[17].

Converting the PD into the PI is essential for constructing topological features. However, it cannot be transformed efficiently if the duration of the 0-dimensional persistent features is infinite. This paper employs a substitution strategy that replaces the infinite duration with the maximum value of the duration of all features. In addition, an auxiliary coordinate is introduced to ensure that the feature on the 0-dimensional PI remains intact after the transformation.

3. Employing Squeeze-and-Excitation (SE)[18] attention mechanism to adaptively learn the importance weights of the fused features.

In the feature fusion process, it is often difficult to effectively discern the importance of different features. In this paper, after feature fusion, the Squeeze-and-Excitation (SE) attention mechanism enables the model to adaptively learn the weighting of the two types of features, further optimizing the feature extraction and representation capabilities.

## 2. Related Work

TDA is a data analysis approach that provides a novel perspective on understanding data. It facilitates gaining deep insights into the intrinsic structure of data and exploring potential non-linear relationships within it. Recently, TDA has been widely applied in machine learning and deep learning fields to improve model decision-making and generalization.

The primary objective of integrating TDA with machine learning or deep learning is to utilize TDA to extract topological features and combine them with other feature extraction methods to enrich the feature space. This combined approach aims to improve the ability to classify and recognize. Muszynski et al.[19] integrated TDA with Support Vector Machine (SVM) to identify features of Atmospheric Rivers (ARs) in climate data to address the uncertainty of weather patterns. This significantly improved the accuracy of atmospheric river pattern recognition. Kei Takahashi et al.[20] initially classified mouse disease image maps using machine learning techniques and then employed TDA and the Non-Homogeneous Poisson process (NHPP) to extract geometric features. They successfully evaluated the structural variances of the vascular structures. Thomas et al[21] combined radionics features with topological features extracted by TDA to predict malignant tumors using ridge regression. The method bridged the gap between the

Risk Stratification System (RSS) in accurately stratifying follicular carcinomas and adenomas.

Nicolas Swenson et al.[22] combined graph representation learning with TDA to construct the PersGNN model. This model captures local and global topological features by analyzing 1- and 2-dimensional PDs of protein structures for highly accurate protein function prediction. Cang et al.[23] proposed the Element-Specific Persistent Homology (ESPH) method to extract 1-dimensional topological features of protein structures. This method unveils the concealed structure-function relationships in biomolecules. Further, they constructed a multi-channel topological neural network (MM-TCNN) by combining ESPH and a deep convolutional neural network to predict protein-ligand binding affinity and protein stability changes after mutation. Kanksha et al.[24] developed the TDA-DL model by extracting topological features from capillary dilatation and skin lesion images. These features were combined with the features extracted from the EfficientNet-B5 model. The result was improved performance that surpassed that of EfficientNet-B5 and other similar models. Faisal Ahmed and Baris Coskunuzer[25] applied TDA in fundus imaging and proposed the Topo-ML and Topo-Net models. It significantly improved classification performance and outperformed similar models by integrating topological features with machine learning and deep learning methods. Chuan-Shen Hu et al.[26] proposed the TopoResNet-101 architecture, which incorporates topological features such as Persistence Curves (PC) and Persistence Statistics (PS) into a ResNet. This model demonstrates better robustness compared to ResNet-101. Hajij M et al.[27] proposed TDA-Net, an integration network of topological structural features and deep learning features. The network contains two branches: a deep branch with the original image as input and a topological branch with a topological feature vector as input. The outputs of both branches are then merged to perform classification for better classification results.

Combining topological features with machine learning algorithms and deep convolutional neural network[28] can effectively extract deep structural information to improve the accuracy and robustness of data feature extraction, classification, analysis, and other tasks[29-33]. These methods have played an active role in data analysis tasks such as neurology, cardiology, hepatology, gene-level and single-cell transcriptomics, drug discovery, evolution, and protein structure analysis[34]. However, current methods mainly rely on topological feature characterization approaches such as PD, PC, PS, and Persistence Landscape (PL)[35]. These approaches have certain limitations in terms of feature representation capabilities. In addition, current methods do not effectively assign weights to different features. In this paper, we employ a more robust PI representation method based on this premise. Our method automatically assigns importance weight to topological and numerical features through the attention mechanism. This approach effectively enhances the capability and effectiveness of feature fusion.

## 3. Methodology

PI is a method of vectorizing a PD to produce a pixel-value matrix while preserving the original PD's interpretability. This process converts the discrete representation of topological features into a continuous and normalized image representation through point density mapping, Gaussian kernel smoothing, and fixed resolution. These processing steps cause the PI to smooth and filter noise, reduce the impact of outliers, and provide a more stable and robust representation of topological features[36].

This paper proposes a network structure called TDA-CNN that combines TDA and CNN. In this approach, the more feature-expressive PI is used as the topological features representation.

These features are merged with the numerical distribution features extracted by the CNN and the feature weights are adjusted by an attention mechanism. This provides a more comprehensive understanding of data features and improves feature learning and representation capabilities.

### 3.1 TDA-CNN Model

This paper discusses our proposed method and model based on image data. Figure 1 shows the workflow of TDA-CNN model. Let the input image be represented as $D \in R^{H \times W \times C}$ (where $H$, $W$, and $C$ denote the image's height, width, and number of channels). It is assumed that the images have been analyzed using TDA and PIs have been generated.

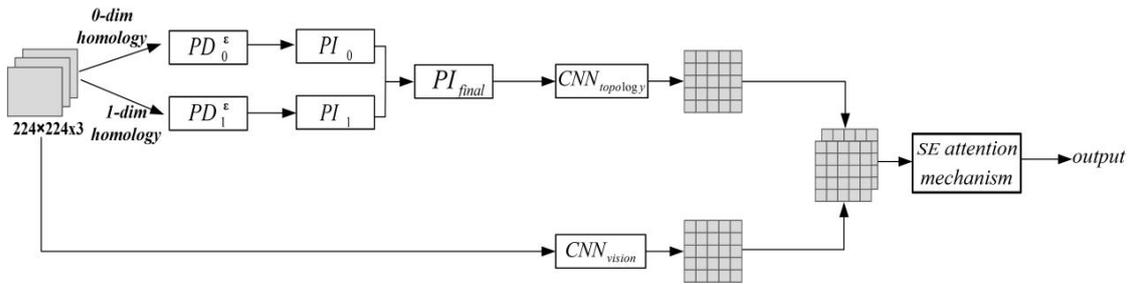

Fig. 1 TDA-CNN model which combines topological data analysis and a convolutional neural network.

1. Numerical features are extracted using $CNN_{vision}$ model[37] to capture local and global features of images $f_{vision}$:

$$f_{vision} = CNN_{vision}(D)$$

2. Using a $CNN_{topology}$ model to process PI obtain $f_{topology}$:

$$f_{topology} = CNN_{topology}(PI)$$

This process must ensure consistency between the dimensions of topological features and visual numerical features, thereby making them compatible for feature fusion.

3. After the feature extraction phase, the numerical features are fused with the topological features through channel splicing to create a more comprehensive fused feature representation $\Psi$:

$$\Psi = f_{vision} \oplus f_{topology}$$

4. The SE attention mechanism is utilized in the fused feature $\Psi$ to enhance the adaptability and learning capability of the model. This allows the TDA-CNN to optimize the quality and efficiency of feature representation by autonomously learning and adjusting the weight distribution between numerical features and topological structure features:

$$\Phi = SE(\Psi)$$

5. The fused features generated by the attention mechanism are fed into the fully connected layer. This step involves mapping the highly abstract feature representation to the category space through the fully connected layer to achieve accurate image classification[36]:

$$output = FC(\Phi)$$

## 3.2 Calculating the Persistent Image（PI）

In the process of calculating PI, there are two issues. Firstly, the 0-dimensional features have zero duration because their birth and death values are equal. These features are clustered on the y-axis when the PD is vectorized into a PI. This renders the PI ineffective. Secondly, the points with infinite duration in the PD are usually isolated or noisy points that cannot be represented in a finite space. These issues above prevent the conversion from PD to PI from functioning correctly. In this paper, the conversion process is improved as follows:

1. Process point clouds of image data to construct a complex shape structure. Then, compute the persistence and extract the topological features. Let the set of extracted $Kth$-dimensional topological features be denoted as $T_K = \{(x_i, y_i)\}_{i=1}^{n}$, where $(x_i, y_i)$ represents the birth time $x_i$ and death time $y_i$ of the $ith$ topological feature.

2. The eligible topological features are filtered by a distance threshold $\varepsilon$ to form a PD, denoted as $PD_K^{\varepsilon} = \{T(x_i, y_i) \mid i = 1, 2, ..., r\}$. Each point in the PD represents a topological feature and its life cycle. The 0- and 1-dimensional $PD_0^\varepsilon$ and $PD_1^\varepsilon$ are computed to visualize the data's topological characteristics and persistence graphically.

For example, Fig.2(b) shows a PD of image (a). This PD, $H_0$ represents a 0-dimensional topological feature indicating connected components or isolated points, and $H_1$ represents 1-dimensional topological features such as rings and holes. The X-coordinate signifies the birth value (the time when the topological feature starts to be generated), and the Y-coordinate signifies the death value (the time when the topological feature ends).

All points are above the diagonal $Y = X$ [38] because topological features persist for some time after they have been generated. The persistence of the feature is represented by the distance from the point to the diagonal. Longer durations indicate more stable topological features, while short durations are considered noisy or unstable features.

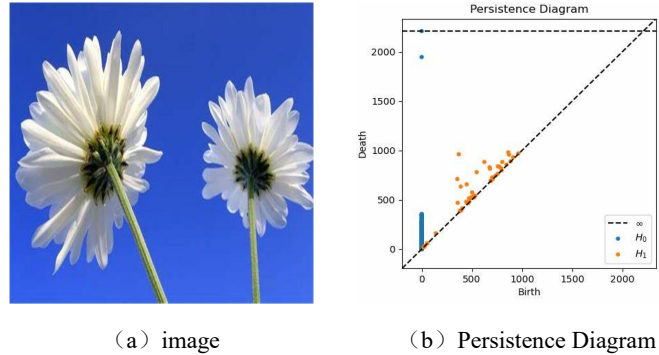

（a）image             （b）Persistence Diagram

Fig. 2 Shows the process of performing topological data analysis on images to obtain a persistent diagram.

3. For 0 and 1-dimensional $PD_0^\varepsilon$、$PD_1^\varepsilon$ the corresponding Birth-Persistence Diagram, BP[41] is constructed by calculating the correlation between the birth time and duration of feature points. The persistence of the feature $i$ is represented as $\pi_i = y_i - x_i$, and the feature is transformed into their birth time and persistence time $(x_i, \pi_i)$.

As shown in Figure 3, for the first issue, the feature $i$ whose duration is infinite is replaced by the maximum persistence time $\pi_{\max}$ in the 0-dimensional BP, and the processed features are denoted as $T^\kappa$. For the second issue, an auxiliary coordinate point $(\tau_x, \tau_y)$ is introduced, where $\tau_x, \tau_y > 0$, $\tau_y > \tau_x$. Through the above operations, the 0-dimensional birth persistence diagram $BP_0^\kappa$ and the 1-dimensional birth persistence diagram $BP_1^\kappa$ are obtained effectively.

4. The PI is generated by transforming the feature point information and their corresponding durations on the 0- and 1-dimensional BPs into pixel value distribution information. The transformation process applies a Gaussian kernel function in the 2-dimensional plane to weigh the feature points and obtain the weighted pixel value distributions.

Centered on each feature $(x_i, \pi_i)$ on the BP, the corresponding topological feature is converted into a PI using a 2-dimensional Gaussian function mapping, which is defined as:

$$\phi(x, y; x_i, \pi_i) = \exp\left(-\frac{(x - x_i)^2 + (y - \pi_i)^2}{2\xi^2}\right) \tag{1}$$

The $PI_{(x,y)}$ is calculated by summing the Gaussian function mentioned above across the image grid.

$$PI_{(x,y)} = \sum_i \phi(x, y; x_i, \pi_i) \tag{2}$$

Where $\xi$ is the standard deviation of the Gaussian kernel, $x$ and $y$ are the coordinates of the image grid. Denote the 0-dimensional persistent image as $PI_0$, and the 1-dimensional persistent image as $PI_1$.

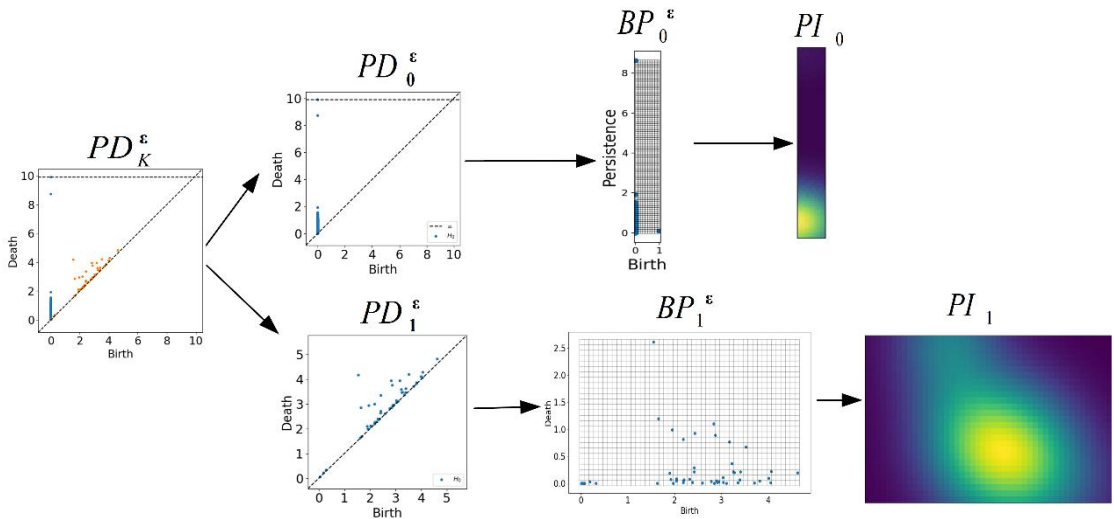

Fig. 3 Illustrates the process of converting images to persistent images.

5. The $PI_0$ and $PI_1$ have different dimensions due to the varying number of features. In

this paper, we utilize the zero-padding method to align the PIs of each dimension to the same size. Then, $PI_0$ and $PI_1$ are stacked using the maximum stacking method to obtain $PI_s$.

$$PI_s = \max(PI_0, PI_1) \tag{3}$$

To enhance the PI's feature representation capability and capture more comprehensive information content, $PI_0$, $PI_1$, and $PI_s$ are stacked to form a three-channel image to ensure that the number of channels is consistent with the original image. As illustrated in Figure 4, this process obtains the final desired PI.

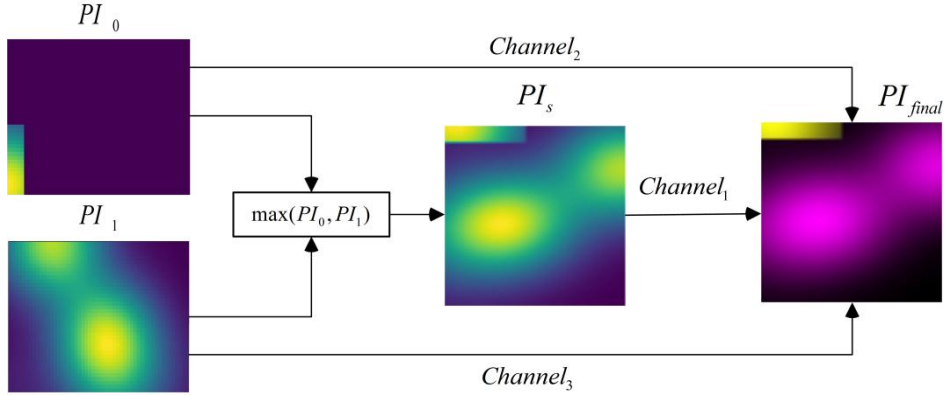

Fig. 4 Fusion based on maximum stacking and channel alignment

### 3.3 Feature Fusion

As shown in Figure 1, $CNN_{vision}$ is used to extract the numerical features of the image and $CNN_{topology}$ is used to extract the topological structural features of the data. The two types of features are fused by channel to create a more comprehensive feature representation, denoted as $\Psi$.

$$\Psi = CNN_{vision}(D) \oplus CNN_{topology}(PI_{final}) \tag{4}$$

A feature map typically has three main dimensions: height, width, and channel. where $\oplus$ is the feature map fusion operation in the channel dimension when the arrays are identical in shape. Suppose there are two feature maps which have the shapes ($H$, $W$, $C_1$) and ($H$, $W$, $C_2$), which are fused in the channel dimension to get the final feature map shape as ( $H$ , $W$ , $C_1 + C_2$ ).

### 3.4 SE Attention Mechanism

The SE attention mechanism is applied to the fused feature $\Psi$ to improve the model's focus on key features and to increase the expressiveness of specific features.

First, global average pooling is performed on the spliced feature $\Psi$ to calculate the importance of each channel.

$$z = \frac{1}{H \times W} \sum_{i=1}^{H} \sum_{j=1}^{W} \Psi(i,j) \tag{5}$$

Where $H$ and $W$ are the height and width of the feature map, respectively. $z \in R^C$ denotes the channel importance vector obtained after pooling, and $C$ is the number of channels.

The channel weights are then learned using fully connected networks.

$$s = \sigma(W_n \bullet \delta(W_1 \bullet Z)) \tag{6}$$

Where $W_1 \in R^{\frac{C}{r} \times C}$ is the weight matrix of the fully connected layer, $\sigma$ and $\delta$ denote the sigmoid activation function and Rectified Linear Unit(ReLU) activation function, respectively. And $r$ is the scaling factor.

Finally, the learned weight vector $s$ is multiplied with the fused features $\Psi$ to adjust the eigenvalues for each channel.

$$\Phi = \Psi(i,j,c) \bullet s(c) \tag{7}$$

Where $\Phi$ represents the feature map adjusted by the SE attention mechanism and $C$ is the channel index. In this way, the final feature $\Phi$ is obtained, which is then fed into the fully connected layer to perform the classification task.

$$output = FC(\Phi) \tag{8}$$

## 4. Experimental Analysis

### 4.1 Datasets and Evaluation Indicators

As shown in Table 1, the experiment utilizes three datasets from different domains. Dataset1 contains image data from natural scenes worldwide, Dataset2 is a gender-categorized dataset, and Dataset3 consists of Chinese characters written in the style of 20 renowned Chinese calligraphers.

The primary evaluation metrics selected were accuracy, precision, recall, and F1 score to assess the model's performance comprehensively.

Table 1 Datasets used in the experiment

|  | Dataset | Number | Classification tasks |
|---|---|---|---|
| Dataset1 | Intel Image Classification | 17046 | six-classification |
| Dataset2 | Gender Classification 200K Images\|CelebA | 202381 | Binary-classification |
| Dataset3 | Chinese Calligraphy Styles by Calligraphers | 105080 | Twenty-classification |

### 4.2 Experimental Design and Parameter Settings

In the topological data analysis phase, the point cloud is processed using the Ripser library to compute persistent homology. This process generates 0- and 1-dimensional PDs and then creates the corresponding PIs to unveil the topological features of the dataset.

Four distinguished convolutional neural network models, including VGG16, ResNet34, GoogleNet (Inception v1), and DenseNet121, are used as the benchmark models. A standard setup with a batch size of 32 was used for model training, and the model performance was gradually optimized through 50 epochs of training. A stochastic Gradient Descent (SGD) optimizer was used to update the model parameters efficiently. The goal was to improve the generalization ability and convergence speed.

### 4.3 Analysis of Results

Table 2 presents the performance comparison between the benchmark models and the TDA-CNN models across three distinct datasets. By comparing the performance metrics of the two models on these datasets, we can gain a deeper understanding of the advantages and performance differences of the TDA-CNN models when dealing with various datasets.

Table 2 Shows comparison between the baseline benchmark models and the TDA-CNN models on three datasets.

| Dataset | Model | accuracy | precision | recall | f1 |
|---------|-------|----------|-----------|--------|-----|
| | VGG16 | 0.8976 | 0.8977 | 0.8976 | 0.8974 |
| | **TDA-VGG16** | **0.9066** | **0.9063** | **0.9066** | **0.9062** |
| | GoogleNet | 0.8869 | 0.8865 | 0.8869 | 0.8866 |
| Dataset1 | **TDA-GoogleNet** | **0.9123** | **0.9118** | **0.9123** | **0.9118** |
| | DenseNet121 | 0.9059 | 0.9056 | 0.9059 | 0.9057 |
| | **TDA-DenseNet121** | **0.9157** | **0.9153** | **0.9157** | **0.9154** |
| | ResNet34 | 0.9026 | 0.9025 | 0.9026 | 0.9024 |
| | **TDA-ResNet34** | **0.9106** | **0.91** | **0.9106** | **0.9102** |
| Dataset2 | VGG16 | 0.9136 | 0.9081 | 0.8852 | 0.8965 |
| | **TDA-VGG16** | **0.9668** | **0.966** | **0.9661** | **0.966** |
| | GoogleNet | 0.9317 | 0.9177 | 0.921 | 0.9194 |
| | **TDA-GoogleNet** | **0.9402** | **0.9272** | **0.9316** | **0.9294** |
| | DenseNet121 | 0.9386 | 0.9401 | 0.9128 | 0.9262 |
| | **TDA-DenseNet121** | **0.9471** | **0.9339** | **0.9414** | **0.9376** |
| | ResNet34 | 0.9368 | 0.9259 | 0.9248 | 0.9254 |
| | **TDA-ResNet34** | **0.9372** | **0.9255** | **0.9259** | **0.9257** |
| Dataset3 | VGG16 | 0.7995 | 0.8006 | 0.7995 | 0.7943 |
| | **TDA-VGG16** | **0.9745** | **0.9744** | **0.9745** | **0.9744** |
| | GoogleNet | 0.8245 | 0.8203 | 0.8245 | 0.8145 |
| | **TDA-GoogleNet** | **0.869** | **0.8665** | **0.869** | **0.8659** |
| | DenseNet121 | 0.8612 | 0.8602 | 0.8612 | 0.8579 |
| | **TDA-DenseNet121** | **0.9323** | **0.9316** | **0.9323** | **0.9316** |
| | ResNet34 | 0.8414 | 0.8384 | 0.8414 | 0.8374 |
| | **TDA-ResNet34** | **0.8685** | **0.8669** | **0.8685** | **0.8661** |

By comparing and analyzing the above experimental results in detail, it is evident that the TDA-CNN models improve accuracy across all datasets. The TDA-VGG16 model in Dataset3 shows the most significant improvement with an increase of 17.50%. The TDA-ResNet34 model in Dataset2 demonstrates the least improvement with an increase of 0.04%. This result is fully verified in the experiments and provides firm support and arguments for the effectiveness and superiority of the TDA-CNN models in multi-domain data processing.

In Figure 5, one image was selected from the test sets of Dataset1 to the probability distribution of each category predicted by the benchmark models and the TDA-CNN models to evaluate the effectiveness of the TDA method in improving classification performance.

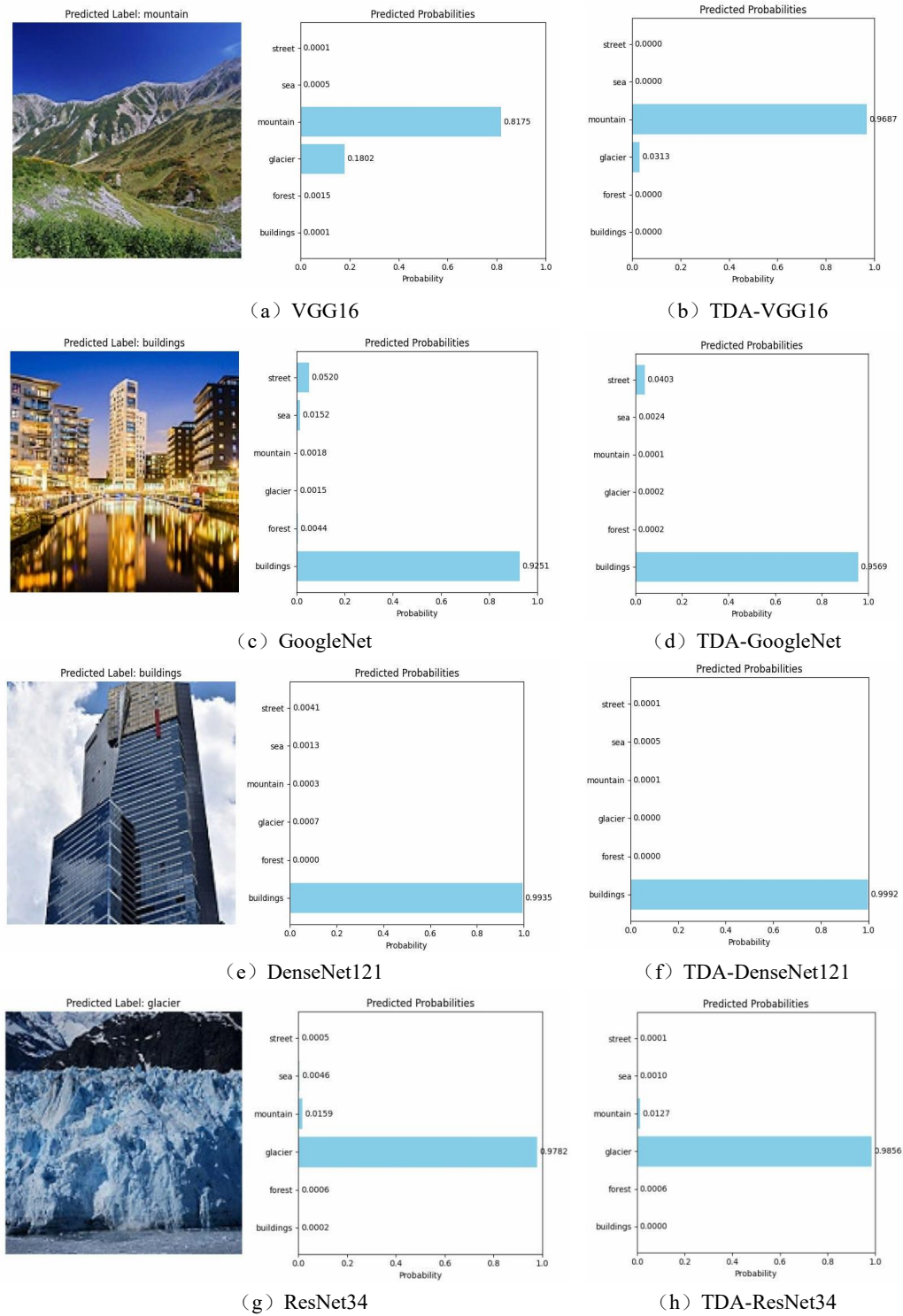

（a）VGG16　　　　　　　　　　　（b）TDA-VGG16

（c）GoogleNet　　　　　　　　　（d）TDA-GoogleNet

（e）DenseNet121　　　　　　　　（f）TDA-DenseNet121

（g）ResNet34　　　　　　　　　　（h）TDA-ResNet34

Fig. 5 Shows the prediction results and probability distribution of each category for the benchmark models and TDA-CNN models on dataset1.

The prediction results and the probability distributions of the categories of the CNN and the

TDA-CNN models are verified. The results show that the prediction of the TDA-CNN models is more accurate. This phenomenon is because the combinion of TDA and CNN can comprehensively capture data feature information to improve classification task accuracy and reliability.

### 4.4 Methodological Analysis

Figure 6 shows the accuracy evolution trend of the benchmark models and TDA-CNN models on the training and validation sets throughout the training process.

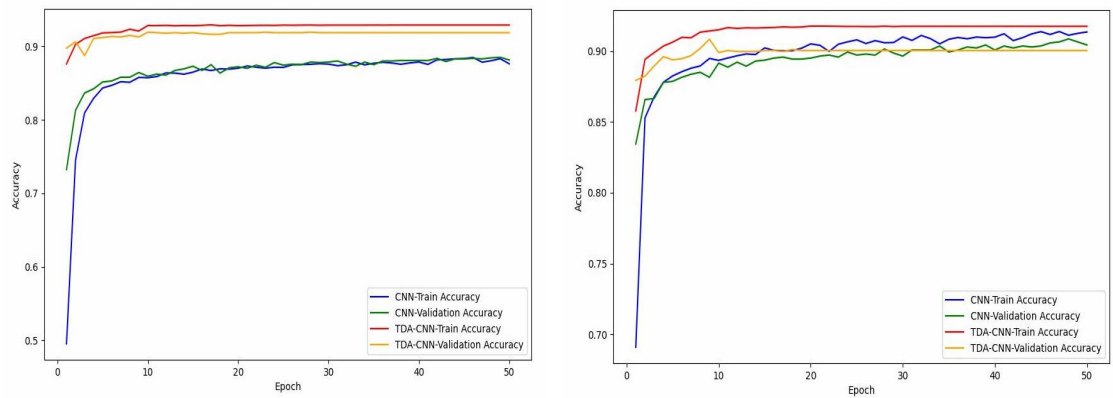

(a)Accuracy comparison between GoogleNet and
TDA-GoogleNet on training and validation sets

(b)Accuracy comparison between DenseNet and
TDA-DenseNet on training and validation sets

Fig. 6 Changes in the accuracy of the benchmark models and TDA-CNN models in each epoch on the training and validation sets.

By comparing the changes in accuracy between the benchmark models and the TDA-CNN model on the training and validation sets, it is evident that the TDA-CNN models outperform the benchmark models. Firstly, the TDA-CNN models demonstrate faster convergence and achieve a high accuracy rate in the early stages of training. Secondly, the accuracy of the TDA-CNN models is significantly higher than that of the benchmark models on both the training and validation sets. This indicates superior learning ability and generalization performance. These results suggest that TDA is capable of mining the intrinsic structural information of the data and effectively capturing features that the benchmark models may overlook. Fusing topological features not only improves the model's understanding of the data but also increases its accuracy in processing complex data. This allows it to make more accurate classifications.

Figure 7 shows the importance of the features extracted by the benchmark models and the TDA-CNN models on classification accuracy.

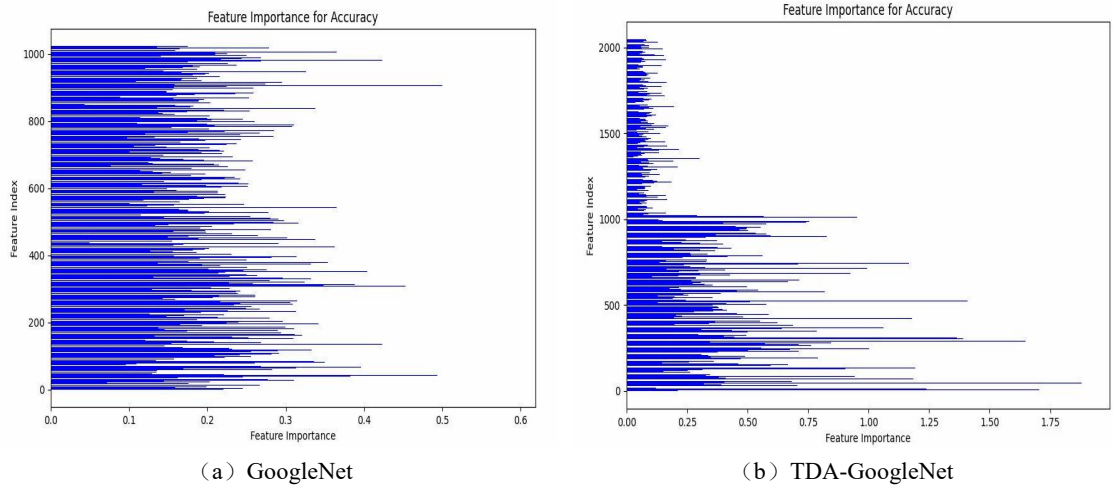

（a）GoogleNet  （b）TDA-GoogleNet

Fig. 7 The Importance in Features of Benchmark Models and TDA-CNN Models for Classification Accuracy.

The significant improvement of the TDA-CNN models in these two aspects is evident when comparing the performance of the benchmark models and TDA-CNN in terms of feature extraction and feature importance. The number of features extracted by the TDA-CNN models increased significantly, and more features had higher importance, up to 1.75. In contrast, the benchmark models extract fewer features, and the feature importance values mainly range between 0.0 and 0.3. This indicates that the TDA-CNN models can extract more valuable and essential features from the data to improve feature representation and learning ability. This is a full demonstration of the effectiveness and superiority of the combinion of TDA and CNN.

Figure 8 shows the visualization results of applying the t-distributed Stochastic Neighbor Embedding (t-SNE) method[39] to reduce the dimensionality of high-dimensional features. The t-SNE method embeds high-dimensional features in a low-dimensional space. This allows their structure and distribution to be represented more intuitively in a two-dimensional graph.

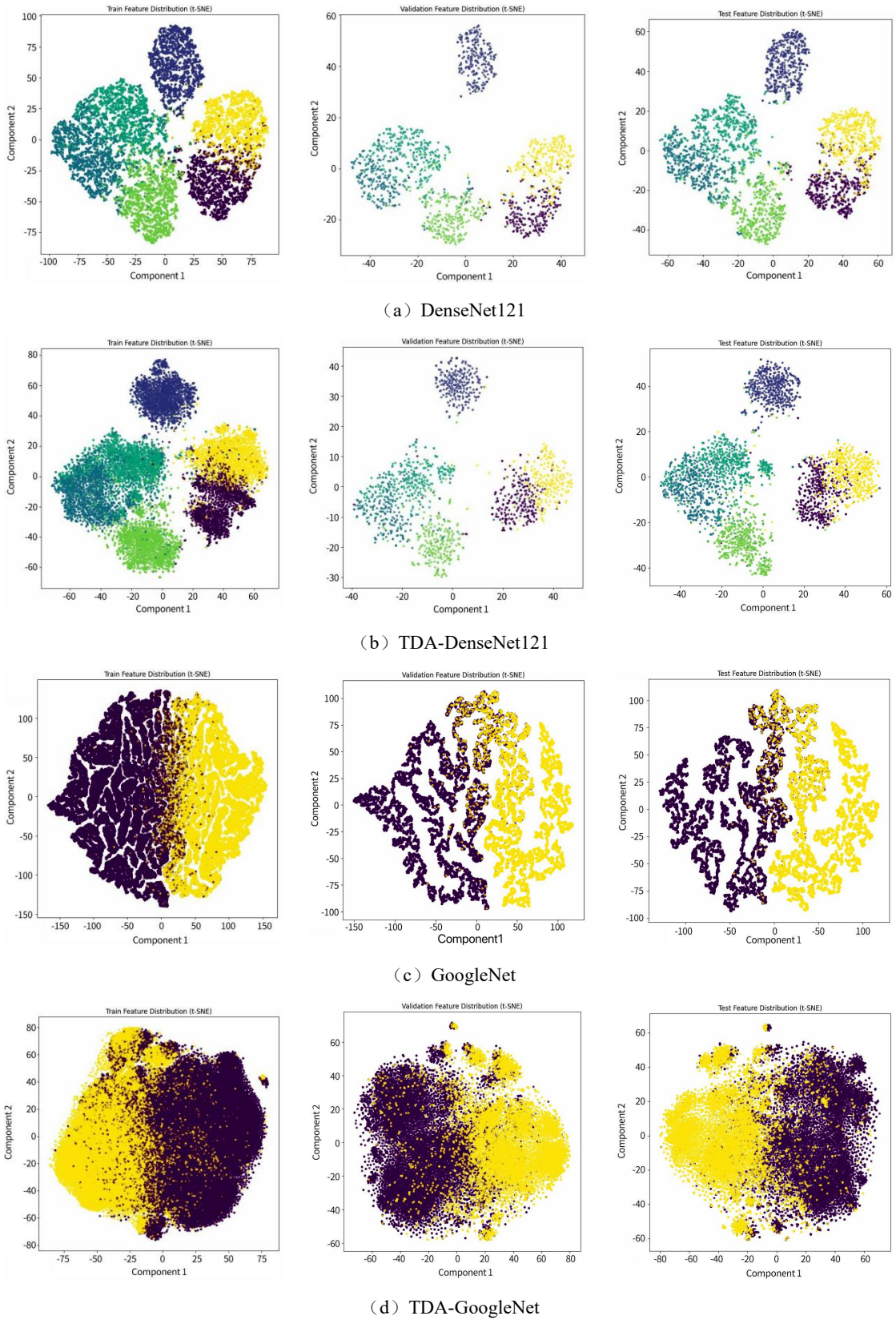

（a）DenseNet121

（b）TDA-DenseNet121

（c）GoogleNet

（d）TDA-GoogleNet

Fig. 8 CNN model and TDA-CNN model visualize the reduced features using t-SNE in dataset1 and dataset2.

By comparing the feature distributions of the benchmark models and the TDA-CNN models using t-SNE for visual analysis. The TDA-CNN demonstrates a more compact cluster structure in feature distribution. This suggests that incorporating TDA improves the model's capacity to extract and analyze features, leading to tighter clustering of features belonging to the same category. Additionally, the separation between the clusters of TDA-CNN models is more pronounced. This indicates that TDA-CNN is more effective in distinguishing and isolating different classes of features.

## 5. Ablation Experiment

The DenseNet121 model was applied to dataset 1 as the baseline model in this ablation experiment, and the various components of the model were fully explored to assess their contribution and impact on the model performance. This experiment has contributed to a deeper understanding of the construction details of the TDA-CNN model, as well as the roles and effects of each component in the image classification task.

Table 3 DenseNet121 ablation experiments on dataset1.

| Experiment group | SE | PI | accuracy | precision | recall | f1 |
|---|---|---|---|---|---|---|
| group1 | × | × | 0.9059 | 0.9056 | 0.9059 | 0.9057 |
| group2 | × | √ | 0.9110 | 0.9107 | 0.9110 | 0.9108 |
| group3 | √ | √ | 0.9157 | 0.9153 | 0.9157 | 0.9154 |

Experimental results show that in Group 1, no additional features are used. In Group 2, the inclusion of PI improves accuracy, precision, recall, and F1 score. This highlights PI's effectiveness in capturing complex data structures and relationships. Furthermore, Group 3, which incorporates the SE attention mechanism along with PI, demonstrates superior performance. It achieves an accuracy improvement of nearly 1% over Group 1 and almost 0.5% over Group 2, with significant enhancements in all metrics. This suggests that adding the SE attention mechanism after fusing CNN and PI enables a more comprehensive representation of the data features, and improves the feature learning ability and classification effectiveness.

## 6. Discussions

TDA is based on algebraic topology theory and provides a powerful method for understanding data structures. This method focuses mainly on the shape and spatial structure of objects. The core concept lies in analyzing the basic characteristics of the object maintained in continuous morphological changes rather than specific metrics or geometric properties. This is known as topological invariance, which facilitates the analysis of the intrinsic structure of the data and provides strong resistance to interference. However, it is precisely because of this highly abstract data structure feature that relying solely on topological structures cannot strictly distinguish the visual differences between objects. As shown in Figure 10, objects such as coffee cups and donuts exhibit similarities in topological structural features but belong to completely different object types. Therefore, there is a need for a fusion of other information for a more accurate distinction and understanding of these objects.

In image classification or visual recognition, it is necessary to comprehensively consider topological features and other image features to maximize the value of topological data analysis. Integrating topological structure features with various information, such as shape, color, texture, etc. It helps to comprehensively consider multiple data aspects, thereby improving the accuracy and credibility of classification. This comprehensive consideration method positively impacts the results of image classification and visual recognition tasks and provides strong support for a deeper understanding and interpretation of data.

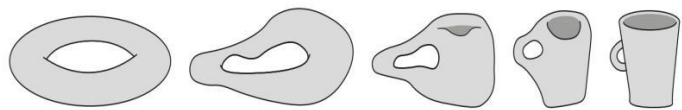

Fig. 9 Coffee cups and donuts are topologically equivalent.

## 7. Conclusion

This paper proposes a fusion method of topological structure features and numerical features, which divides feature extraction into the CNN channel for extracting numerical features and the TDA channel for extracting topological features. The SE attention mechanism enables the model to adaptively learn the weight allocation of feature points in both feature forms, further optimizing the model's feature extraction and representation capabilities. This paper also adopts a replacement strategy and auxiliary coordinate points to solve the issues of zero and infinite duration in feature representation conversion that prevents normal conversion.

The effectiveness of this method has been validated on various datasets and network architectures. Experimental results on datasets such as Intel Image Classification, Gender Classification 200K Images | CelebA, and Chinese Calligraphy Styles by Calligraphers have shown that TDA-CNN improves the performance by 17.5%, 7.11%, and 4.45% on VGG16, DenseNet121, and GoogleNet networks, respectively. The experiment shows that integrating the advanced abstract features of TDA with deep learning ability can effectively improve the performance of feature extraction, classification, analysis, and other tasks.

At present, in addition to investigating the use of topological features to enhance feature representation capabilities, we also consider to use them as physical information to guide generative models so that the generated data conform to specific topological structure requirements. Furthermore, we explore ways to reduce the additional computational burden of TDA guidance.

## Acknowledgments

This research was supported by the Academic Research Projects of Beijing Union University (ZK10202403). We want to express our sincere gratitude to all those who provided valuable insights and assistance during this study. Additionally, we acknowledge the contributions of our research team members for their dedication and hard work.